\documentclass[journal]{IEEEtran}
\usepackage{color}
\usepackage{amsmath}
\usepackage{amssymb}

\usepackage{bbm} 

\usepackage{url}
\usepackage{cite}
\usepackage[breaklinks=true,bookmarks=false]{hyperref}
\usepackage{booktabs}
\usepackage{multicol}
\usepackage{multirow}

\usepackage{bbding}
\usepackage{pifont}

\ifCLASSINFOpdf
   \usepackage[pdftex]{graphicx}
\else
\fi

\begin{document}

\title{Pose Neural Fabrics Search}

%
%
%

\author{Sen Yang,~
	Wankou Yang,~\IEEEmembership{Member,~IEEE}
	and~Zhen~Cui,~\IEEEmembership{Member,~IEEE}
	
	\thanks{Sen Yang and Wankou Yang are with the School of Automation of Southeast University, Nanjing 210096, China. E-mail: yangsenius@seu.edu.cn, wkyang@seu.edu.cn. (Corresponding author: Wankou Yang.)}
	\thanks{Zhen Cui is with the School of Computer Science and Engineering, Nanjing University of Science and Technology, Nanjing, China. E-mail: zhen.cui@njust.edu.cn.}
}

\maketitle

\begin{abstract}

  Neural Architecture Search (NAS) technologies have emerged in many domains to jointly learn the architectures and weights of the neural network. However, most existing NAS works claim they are task-specific and focus only on optimizing a single architecture to replace a human-designed neural network, in fact, their search processes are almost independent of domain knowledge of the tasks. In this paper, we propose Pose Neural Fabrics Search (PoseNFS). We explore a new solution for NAS and human pose estimation task: part-specific neural architecture search, which can be seen as a variant of multi-task learning. Firstly, we design a new neural architecture search space, Cell-based Neural Fabric (CNF), to learn micro as well as macro neural architecture using a differentiable search strategy. Then, we view locating human keypoints as multiple disentangled prediction sub-tasks, and then use prior knowledge of body structure as guidance to search for multiple part-specific neural architectures for different human parts. After search, all these part-specific CNFs have distinct micro and macro architecture parameters. The results show that such knowledge-guided NAS-based architectures have obvious performance improvements to a hand-designed part-based baseline model. The experiments on MPII and MS-COCO datasets demonstrate that PoseNFS\footnote{Code is available at \url{https://github.com/yangsenius/PoseNFS}} can achieve comparable performance to some efficient and state-of-the-art methods.
\end{abstract}

\begin{IEEEkeywords}
Human pose estimation, neural architecture search, cell-based neural fabrics, micro and macro search space, prior knowledge, part-specific NAS, vector representation
\end{IEEEkeywords}

%
\IEEEpeerreviewmaketitle


\section{Introduction}

\IEEEPARstart{N}{eural} Architecture Search (NAS), the process of jointly learning the architecture and weights of the neural network \cite{liu2018darts,Zoph2016NeuralAS, Ghiasi_2019_CVPR,chen2018searching,zoph2018learning,elsken2018neural,real2018regularized,Xie_2019_ICCV,liu2019auto}, can play a potential role at designing efficient network architectures automatically. Current NAS methods mainly take image classification as basic task, and only search for a micro “cell” to build a chain-like structure. However, when applying NAS to dense (pixel-wise) prediction tasks such as semantic segmentation and human pose estimation, the micro search space is no longer able to generate more complex architectures. Therefore, it become a necessity to artificially design a macro search space allowing identifying a hierarchical structure upon cells for these tasks. In addition, most existing NAS works such as~\cite{liu2018darts,Ghiasi_2019_CVPR,liu2019auto,xie2018snas, hu2020dsnas} optimize a single architecture in the search space, and finally obtain a so-called task-specific architecture to replace a human-designed architecture in the pipeline. Such practice, in fact, decouples the automating architecture engineering from the characteristics of tasks, failing to take advantage of significant domain knowledge to guide the search process and achieve the expected targets.
\begin{figure}
	\begin{center}
		\includegraphics[width=0.45\textwidth]{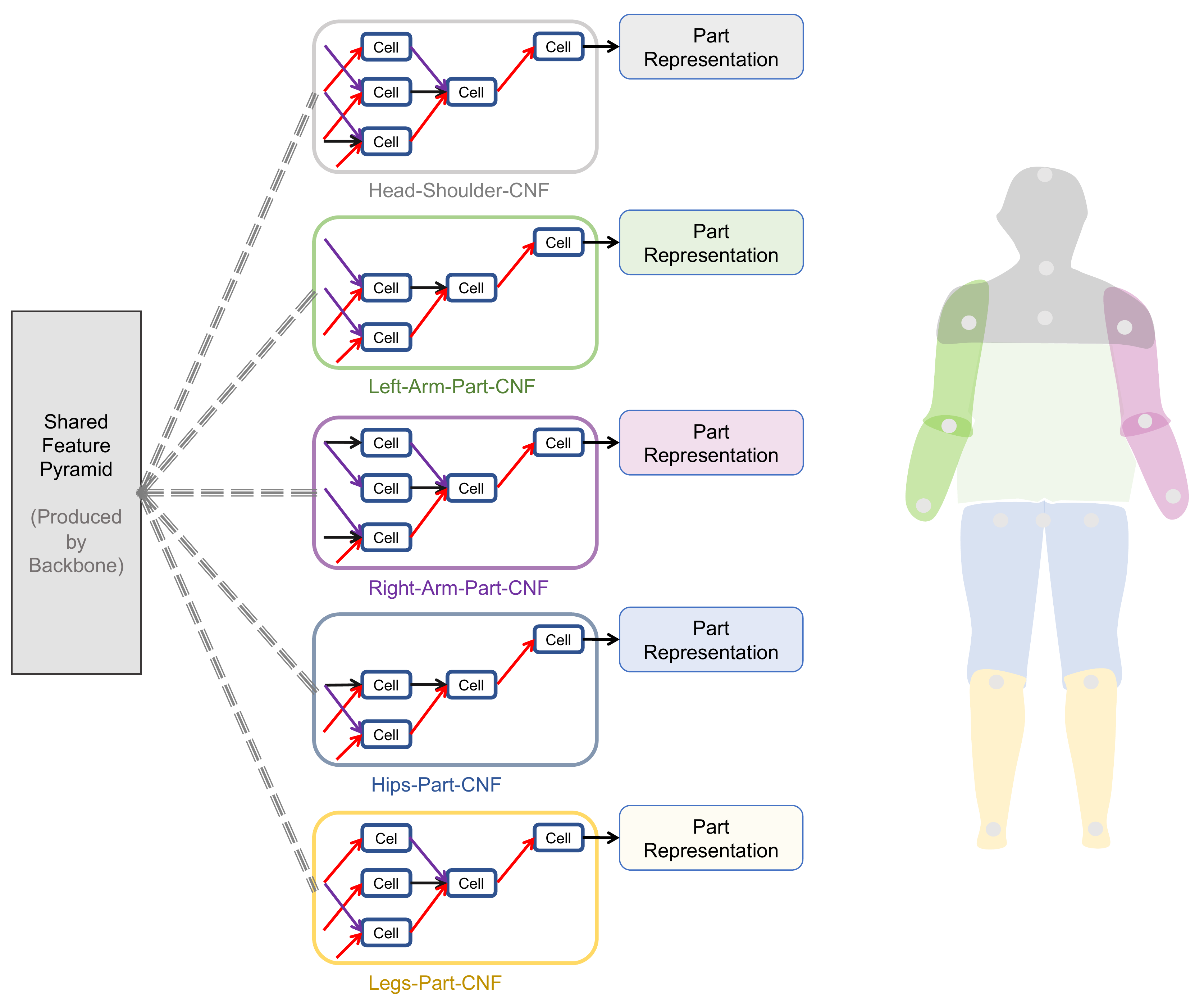} 
	\end{center}
	\caption{ A schematic diagram of searched architectures: five part-specific CNFs (Cell-based Neural Fabrics) are associated with different body parts according to the prior body structure. They take as input the shared feature pyramid produced from a common CNN backbone or a derived CNF. The weighted operations combinations in the cells are also distinct in the part-specific CNFs.}
	\label{f_show}
\end{figure}

For human pose estimation task, we argue that the prior knowledge of human body structure can help discover specific neural architectures in a part-based and automatic way. Thus, we propose a new paradigm: part specific neural architecture search, to search multiple neural architectures for different human parts with guide of prior knowledge, as shown in Fig.~\ref{f_show}.

\begin{figure}
 	
	\begin{center}
		\includegraphics[height=0.25\textwidth]{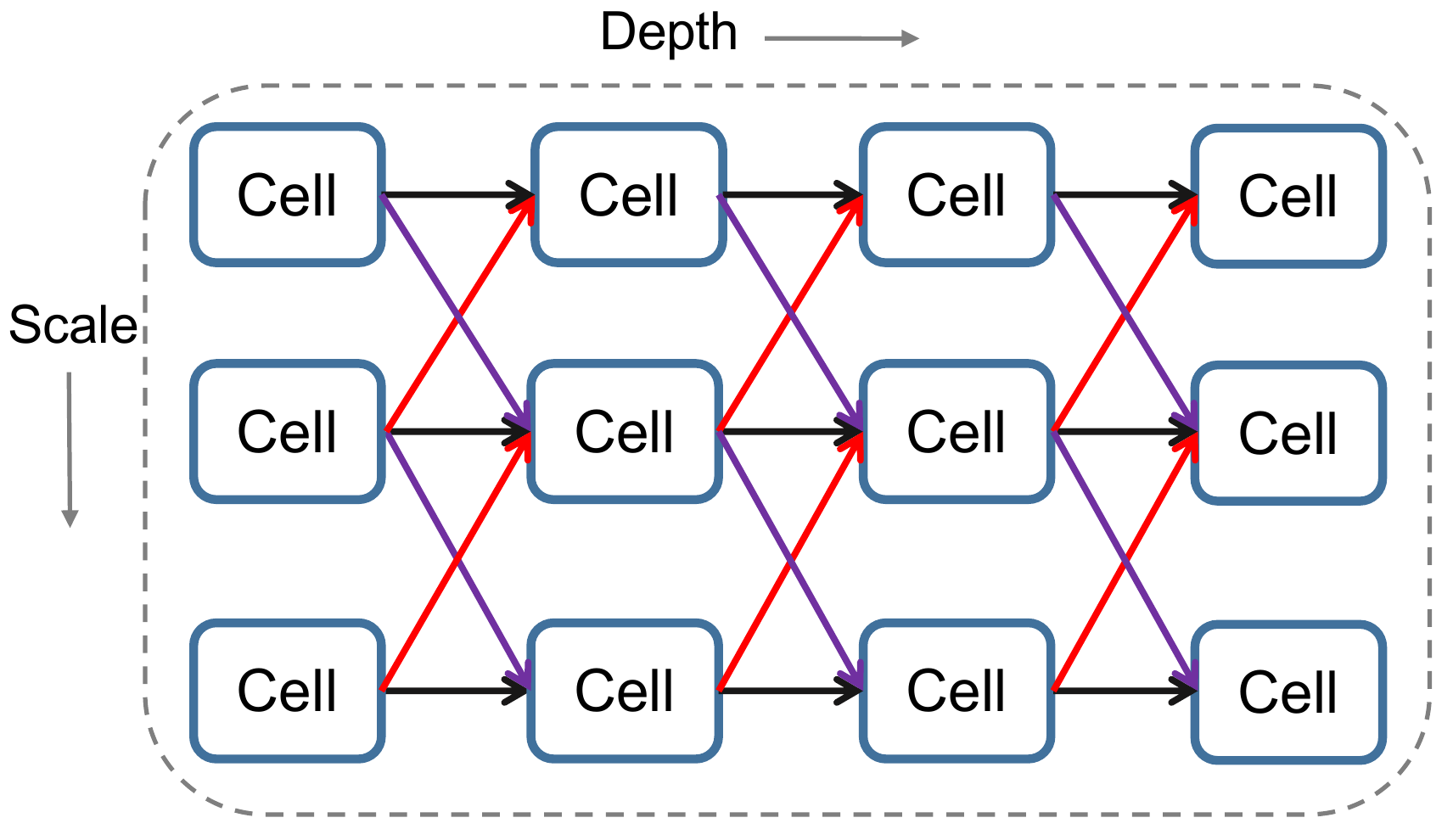} 
	\end{center}
	\caption{A schematic map of the structure of Cell-based Neural Fabric (CNF). The neural fabric is woven by cells in different scales and layer depths. Black arrow represents identity transformation; purple arrow represents reducing spatial size and doubling the channels of feature maps; red arrow represents increasing spatial size and halving the channels of feature maps. }
\label{cnf_}
\end{figure}

Naturally, the first step to introduce NAS into human pose estimation is to construct an architecture search space that can identify multi-scale, stacked or cascaded neural network. To this end, we propose a general parameterized Cell-based Neural Fabric (CNF), a scalable topology structure to encode micro and macro architecture parameters into cells. The discrete search space is relaxed into continuous search space to make it searchable by gradient descent. This design is motivated by Convolutional Neural Fabrics \cite{saxena2016convolutional} and DARTS \cite{liu2018darts}, it can be described as a neural fabric architecture woven by cells, as shown in Fig.~\ref{cnf_}.

In addition, there exists an inconsistency gap between the derived child network (sub-architecture) and the converged parent network (super-network) in DARTS \cite{liu2018darts}. Many works attempt to eliminate this bias such as SNAS-series works \cite{xie2018snas, hu2020dsnas}. In our work, we avoid it in a direct and simple way. We do not re-discretize the continuous architecture after searching, which means that all operations are densely preserved with macro and micro architecture parameters in both searching and evaluation stage. In order to verify such a setting, we test the performances of architectures randomly sampled from the continuous parameter space. Furthermore, we explore a simple yet effective method, gradient-based synchronous optimization, as the major search strategy to reduce the cost of time and computational budgets for the search process.

Designing the neural search space or search strategy is just the beginning for NAS, we believe domain knowledge can help NAS go further. For human pose estimation task, the special human body structure information is significant prior knowledge to be exploited. Current ConvNet-based human pose estimation methods~\cite{wei2016convolutional, cao2017realtime, papandreou2017towards, fang2017rmpe, yang2017learning, chen2018cascaded, xiao2018simple, Sun_2019_CVPR} usually use FCN-like architecture to predict keypoints heatmaps. The global spatial relationship between all keypoints is implicitly learned by shared convolutional blocks and a single linear prediction head. Nevertheless, such learning methods may ignore the explicit knowledge: the spatial distributions of keypoints might be highly correlated in the short-range while weakly in the long-range. Tang et al.~\cite{Tang_2019_CVPR} have found that some pairs of body parts are weakly related by analyzing their mutual information, and a sharing mechanism using the same network to learn shared feature for all keypoints may cause negative transfer between uncorrelated parts. They therefore view human pose estimation as homogeneous multi-task learning (MTL)~\cite{caruana1998multitask,ruder2017an}, and design a structure-shared part-based branching network (PBN) to estimate related parts. Here we ask: since human pose estimation can be viewed as multiple prediction tasks, the internal characteristics of these tasks would be different, are structure-shared branching networks really better for multiple sub-tasks? Whether can NAS search for more specific neural architectures to localize these disentangled human parts?

\begin{figure*}[h]
	\begin{center}
		\includegraphics[height=6cm, width=18cm]{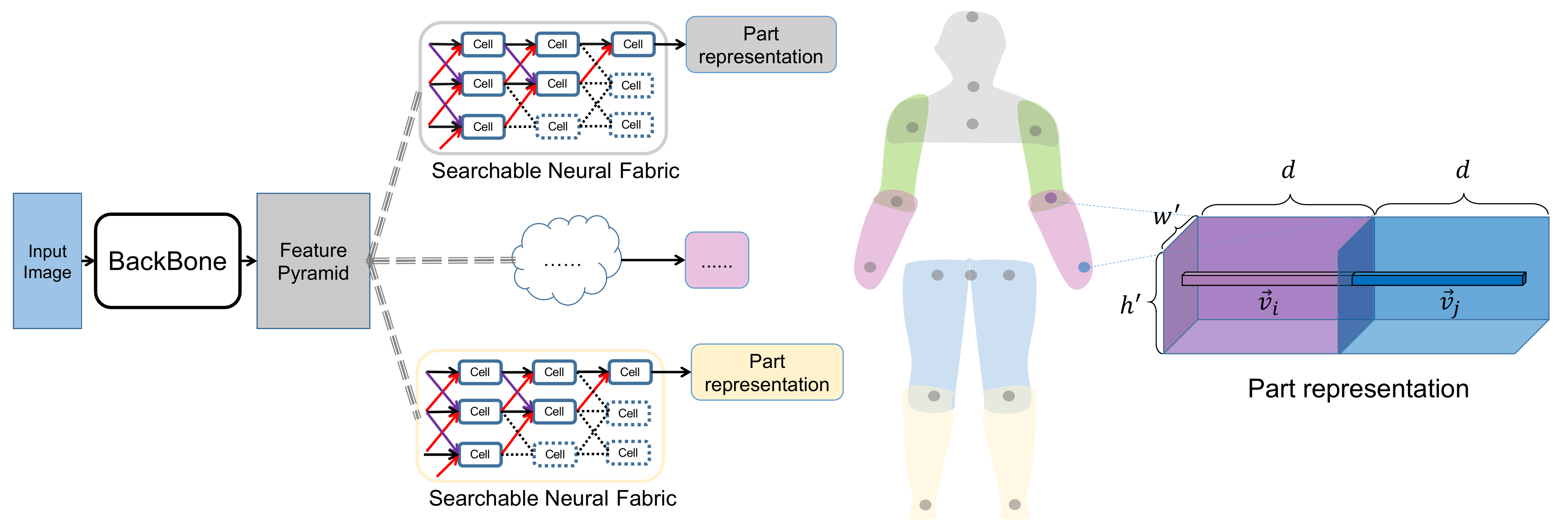}
	\end{center}
	\caption{Pose neural fabrics search framework. \textbf{Left:} Given an input image, feature pyramid representing multi-scale feature maps will be produced from backbone network. Then there will be $P$ CNFs receiving the same feature pyramid and predicting $P$ part representations, here two CNFs shown for simplification. The final cell in the highest scale produces the part representation. Dashed lines mean unused connections and cells. \textbf{Mid:} The whole body is divided into multiple parts associated with keypoints.  \textbf{Right:} For instance, right lower arm part representation is associated with the wrist and elbow keypoints. $\vec{\mathbf{v}}_i$ and $\vec{\mathbf{v}}_j$ mean two $d$-dim vectors respectively for wrist and elbow keypoints at location $(x,y)$ of part representation feature map, and it's shape is $2 \times w^{'} \times h^{'} \times d $.}
	\label{framework}
\end{figure*}

As a matter of fact, such problems can be converted into searching multiple task-specific neural architectures for multiple sub-tasks. There also has been a new trend to explore Multi-task Neural Architecture Search such as~\cite{liang2018evolutionary,newell2019feature,gao2020mtl}, due to that MTL could improve generalization by leveraging the domain-specific information contained in the training signals of related tasks~\cite{caruana1998multitask}. In this work, provided with the proposed search space, we can search multiple architectures to achieve multiple learning objectives in a single shot, by exploiting simple prior knowledge as guidance.  As a result, part-specific CNFs with distinct micro and macro architecture parameters are achieved to adapt to the characteristics of different sub-tasks. As shown in Fig.~\ref{framework}, highly correlated keypoints are grouped into the corresponding part representations that are predicted by searchable CNFs (multi-head predictions). Then each location of keypoint is inferred by the associated part representation. Such NAS-based part-specific architectures extend the carefully designed message passing mechanism, hand-designed branch or information fusion module~\cite{Chu_2016_CVPR,Tang_2019_CVPR, li2014heterogeneous} to fuse feature information between related joints. Besides, we replace scalar value by a vector in each pixel position of heatmaps, because the scalar value representing the existing probability of keypoints is still inadequate to encode local feature. We use the length ($\ell_{2}$ norm) of the vector to represent the existing score of the keypoint and vector space to capture more local component information of the body part. In summary, our main contributions are:

\begin{itemize}
	\item We propose a novel micro and macro architecture search space: parameterized Cell-based Neural Fabric (CNF).
	\item With simple prior knowledge as guidance, our method automatically searches part-specific neural architectures to localize disentangled body parts, which extends the traditional part-based methods.
	\item Such part-specific neural architecture search can be seen as a variant of multi-task learning. It is a novel NAS paradigm as Multi-Task Neural Architecture Search for human pose estimation task.
	\item The experiments show that NAS-based part-specific architectures have obvious performance improvements to a hand-designed part-based baseline model. And our models achieve comparable performances to some efficient and state-of-the-art methods.

\end{itemize}

\section{Related Work}

\paragraph{Neural Architecture Search} Our work is motivated by convolutional neural fabrics \cite{saxena2016convolutional} and neural architecture search methods \cite{liu2018darts,chen2018searching,liu2019auto,Xie_2019_ICCV}. Liu et al. \cite{liu2018darts} propose the continuous relaxation for architecture representation to search architectures using differentiable strategy. Chen et al. \cite{chen2018searching} adopt random search to search Dense Prediction Cell architecture for dense image prediction. Ghiasi et al. \cite{Ghiasi_2019_CVPR} use Reinforce Learning to explore more connection possibilities in different scales of feature pyramid network. Xie et al. \cite{Xie_2019_ICCV} explore randomly wired networks for image classification and proposed the concept of \emph{network generator}.  Liu et al. \cite{liu2019auto} propose Auto-DeepLab, searching for a hierarchical neural network as backbone to replace original human-designed network in a common pipeline (DeepLab) for semantic segmentation. It aims to search a cell structure and a better path in multiple branches. In contrast, our architecture space construction method is one-step, unlike the two-step construction scheme for architecture search space in AutoDeepLab, the whole architecture at macro and micro level is totally parameterized as each cell's parameters and not pruned into sparsely connected architecture. Our work also can be viewed as Multi-task Neural Architecture Search. There are similar works like~\cite{liang2018evolutionary, newell2019feature, gao2020mtl}.

\paragraph{Human pose estimation and part-based model}
Top-down \cite{wei2016convolutional,xiao2018simple,Sun_2019_CVPR,chen2018cascaded,fang2017rmpe,he2017mask} and bottom-up \cite{cao2017realtime,insafutdinov2016eccv,insafutdinov2017arttrack,kocabas2018multiposenet,newell2017associative} human pose estimation approaches have been proven extremely successful in learning global or long-range dependencies relationships of body pose. However, parts occlusions, viewpoint variations, crowed scene, and background interference etc. are still tough problems. Compositional structure models or part-based models \cite{belagiannis2017recurrent,tang2018deeply,sun2017compositional,bienenstock1997compositionality,felzenszwalb2008discriminatively,andriluka2009pictorial,felzenszwalb2005pictorial,park2017attribute,sun2011articulated,7226826,Tang_2019_CVPR,8482293} attempt to overcome aforementioned problems by representing the human body as a hierarchy of parts and subparts. Early part-based models \cite{sun2011articulated,yang2012articulated},  pictorial structure models \cite{felzenszwalb2005pictorial, felzenszwalb2008discriminatively, andriluka2009pictorial}  and some of the current models \cite{tompson2014joint, Chu_2016_CVPR, chen2017adversarial, ke2018multi, tang2018deeply, Tang_2019_CVPR} are usually based on hand-crafted feature (e.g. HOG, SIFT), hand-designed convolutional neural networks or hand-designed information fusion modules. Our method also exploits the compositionality of body pose to separately predict related keypoints, but further develops it by employing NAS to learn distinct CNFs for different keypoints groups.

\paragraph{Vector Representation}
The vector in pixel method is motivated by embedding and vector representation methods \cite{newell2017associative,Papandreou2018PersonLabPP,cao2017realtime,papandreou2017towards, e2018matrix,sabour2017dynamic}. Newell et al. \cite{newell2017associative} propose associative embedding to group body keypoints. Papandreou et al. \cite{Papandreou2018PersonLabPP} use geometric embedding representation to predict offset vectors of keypoints. Cao et al. \cite{cao2017realtime} use part affinity vector field to supervise the part prediction. In addition, Hinton et al. \cite{e2018matrix} use matrix with extra scalar to represent an entity. Sabour et al. \cite{sabour2017dynamic} propose Activity Vector, its length can represent existing probability and its orientation represents the instantiation parameters. In this work, we view each type of keypoint as a type of entity in the image and use activity vectors to locate keypoints to estimate human pose.

\section{Pose Neural Fabrics Search}

\subsection{Overview}

\begin{figure}
	
	\centering
	\includegraphics[height=2.6cm, width=8cm]{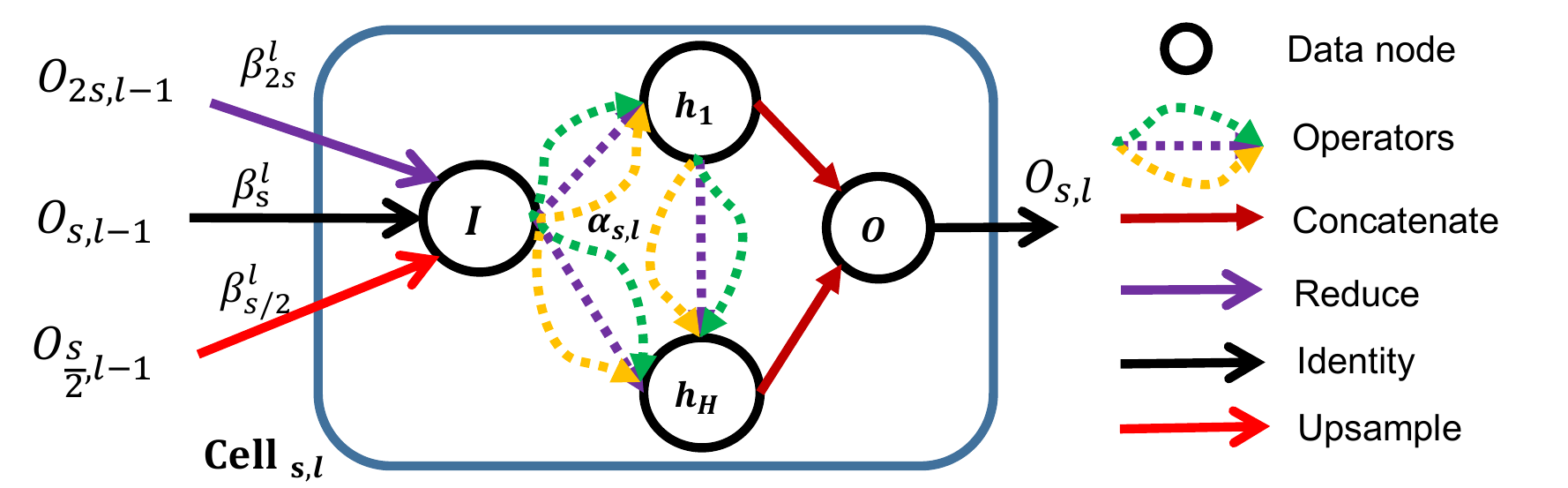}
	\caption{An overview of inner structure of a cell in scale $s$ and $l$ layer. To simply present the inner structure of cell, we set the number of hidden nodes $H$ as 2. Actually, its number of hidden nodes can be 1 or more than 2, each hidden node $h_H$ is densely connected from its previous nodes $\left\lbrace h_0,h_1,...,h_H \right\rbrace $.   }
	\label{cell}
\end{figure}

\begin{figure*}
	
	\centering
	\includegraphics[height=4.8cm, width=16.6cm]{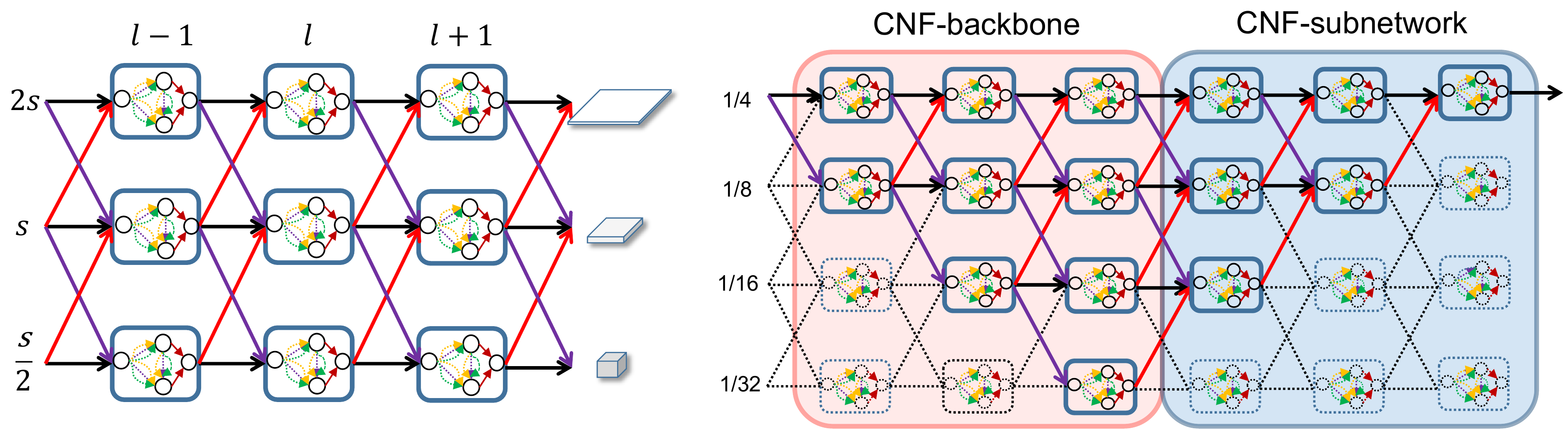}
	\caption{An overview of the CNF neural search space. \textbf{Left:} The homogeneous local connectivity between cells in a neural fabric. \textbf{Right}: Examples of constructing a CNF-backbone (red box) or a CNF-subnetwork (blue box) from CNF. Dashed lines mean unused connections and cells.}
	\label{fabric}
\end{figure*}

Based on top-down method, 2D human pose estimation aims to locate all $K$ keypoints coordinates of body joints(e.g., shoulder, wrist, knee, etc) in a single pose. Let $S=\left \{(x_i,y_i)|x,y\in \mathbb{R^+}, i=1,2,...,K  \right \} $ denote the set consisting of all keypoints coordinates. Considering human body skeleton structures, we disentangle the whole body pose into $P$ part representations as $P$ prediction sub-tasks. $P$ subnetworks are constructed from CNFs sharing backbone to  predict related keypoints subset $s$ ($s\subseteq S$) whose element is associated with the corresponding part. The vector in pixel method is introduced to capture keypoint's feature in a relaxed vector space and the prediction of specified keypoint's location is determined by the location of vector $\vec{\mathbf{v}}$ whose length is the largest. The entire framework is shown in Fig. \ref{framework}.

In Section~\ref{arch}, \ref{subnet}, \ref{optimization}, we demonstrate how to design and employ parameterized CNFs as the choice of subnetworks and backbone. Then, we describe how to randomly sample architectures and optimize the models by synchronous optimization. In Section \ref{prior}, we describe what prior structures of human body are employed to guide neural architecture search. In Section \ref{vector in pixel}, we demonstrate how to utilize the vector representation to estimate keypoints' locations.

\subsection{Neural Architecture Search Space}

\label{arch}
\textbf{Micro structure.} Cell is a repeatable unit across different layers and scales of the whole architecture. Illustrated in Fig.~\ref{cell}, it receives outputs from previous layer's cells as its single input node $I$ and it has $H$ nodes as it's hidden states. Each hidden node $h_j$ is connected by a directed edge with each element of candidate nodes set $\left \{ h_0,h_1,h_2,...,h_{j-1}\right \}$ $\left ( h_0=I,j=1,2,...,H\right)$. Continuous Relaxation \cite{liu2018darts} method is adopted to represent each directed edge with mixed operations. For each $h_j$ is computed by:
\begin{equation}
h_j=\sum_{i=0}^{j-1}\sum_{o \in \mathcal{O}}\frac{\exp \left(\alpha_{o}^{(i, j)}\right)}{\sum_{o^{\prime} \in \mathcal{O}} \exp \left(\alpha_{o^{\prime}}^{(i, j)}\right)} o(h_i),
\end{equation}
where $\mathcal{O}$ is the set of candidate operations. We choose 6 types of basic operations $o$, consisting of: zero, skip connection, $3\times3$ depthwise separable convolution, $3\times3$ dilated convolution with 2 rate, $3\times3$ average pooling, $3\times3$ max pooling. $\alpha_{o}^{(i, j)}$ means the associated weight for each operation $o\in \mathcal{O}$ in edge $h_i \rightarrow h_j$. For a specified $Cell_{s,l}$ in scale $s$ and layer $l$ in CNF, the continuous search space at micro level is: $\alpha_{s,l}=\left\{\alpha_{o}^{(i, j)}|\forall o\in\mathcal{O},\forall j\in\left\lbrace 1,2,...,H\right\rbrace ,\forall i\in\left\lbrace 0,1,...,j-1 \right\rbrace \right\}$, $\alpha_{s,l}\in \mathbb{R}^{\left\|\mathcal{O}\right\|\times\frac{H \left( H+1\right)}{2}}$. All hidden nodes $\left \{ h_1,h_2,...,h_H\right \}$ are concatenated together and reduced in channels by a $1\times1$ conv to achieve an independent output node $O_{s,l}$.

\textbf{Macro Structure.} For $Cell_{s,l}$, it receives the sum of outputs from cells in the previous layer: $O_{2s,l-1}$ , $O_{\frac{s}{2},l-1}$ and $O_{s,l-1}$. They are associated with macro architecture parameters $\beta_{s,l}=\left\{\beta_{2s}^l,\beta_{s}^l,\beta_{\frac{s}{2}}^l\right\}\in \mathbb{R}^3$, which are normalized to control different information reception level from previous cells in different scales. All $\beta_{s,l}$ in all cells of the fabric construct the macro continuous search space. For each $Cell_{s,l}$, its $I$ is computed by:
\begin{equation}
I=h_0=\sum_{(O,\beta)\in Z_{s,l} }\frac{\exp \left(\beta\right)}{\sum_{\beta^{\prime} \in \beta_{s,l}} \exp \left(\beta^{\prime}\right)} \mathcal{T}(O),
\end{equation}
where $Z_{s,l}=\left\{(O_{2s,l-1},\beta_{2s}^l),(O_{s,l-1},\beta_{s}^l),(O_{\frac{s}{2},l-1},\beta_{\frac{s}{2}}^l)\right\}$ and $\mathcal{T}(\cdot)$ is scale transformation operation. In particular, $\mathcal{T}(O_{2s,l-1})$ is downsampling operation via Conv-BN-ReLU mode with 2 stride (meanwhile doubling the channels of data node), $\mathcal{T}(O_{\frac{s}{2},l-1})$ is upsampling operation via bilinear interpolation (meanwhile halving the channels of data node by 1$\times$1 conv) and $\mathcal{T}(O_{s,l-1})$ means identity transformation.

\textbf{Parametric Form.} In summary, we parameterize the form of cell in the $l$-th layer and $s$-th scale of CNF in such a pattern:
\begin{equation}
O_{s,l} =\mbox{Cell}_{s,l }\left(O_{2s,l-1},O_{s,l-1},O_{\frac{s}{2},l-1};w_{s,l},\alpha_{s,l},\beta_{s,l}\right),
\label{con}
\end{equation}
where $w_{s,l}$ represents the weights of all operations in each cell, $\alpha_{s,l}$ and $\beta_{s,l}$ encode architecture search space inside the cell. The hyperparameter of each cell is $\theta_{s,l}=\left(H,C,\mathcal{O}\right)$. $H$ is the number of hidden nodes in each cell. $C$ is the channel factor for each node to control the model capacity, i.e., the number of channels of its data node is $C\times\frac{1}{s}$. $\mathcal{O}$ is the set of candidate operations. This form also can be seen as a scalable topology structure or a network generator \cite{Xie_2019_ICCV}, which can map from a parameter space $ \Theta $ to a space of neural network architectures $\mathcal{N}$, formulated as $g:\Theta \longmapsto\mathcal{N}$.

\subsection{Constructing Subnetworks or Backbone}
\label{subnet}
Benefit from its local homogeneous connectivity pattern as shown in the left of Fig.~\ref{fabric}, cell-based fabric is very flexible and easy to extend into different layers and scales for high and low resource use cases. It is determined by a group of hyperparameters $\Theta = (L, \left\{1/1,1/2,...,1/2^b\right\}, H, C, \mathcal{O})$ where $L$ is total layers and $1/2^b$ is the smallest scale. Illustrated in Fig.~\ref{fabric}, fabric backbone can be constructed by reserving the first $m$ layers and discarding the latter $L-m$ layers to produce feature pyramid in multiple scales. Likewise, fabric subnetwork can be constructed by reserving the latter $n$ layers and discarding the first $L-n$ layers to receive feature pyramid from backbone. Note that our backbone is not restricted to the proposed architecture, and we do not use $3\times3$ average pooling and $3\times3$ max pooling as candidate operations for subnetworks as they are empirically more suitable for extracting low-level visual feature rather high-level feature.

Following common practice for pose estimation, the smallest scale is set to 1/32. We use a two-layer convolutional stem structure to firstly reduce the resolution to 1/4 scale, and consecutively weave the whole CNF \footnote{For cells in first layer, they only receive the stem's output. And for cells in 1/32 scale or 1/4 scale of its current layer, it may have only have two outputs from previous cells. In this case, we will copy one of candidate inputs.}. In order to achieve a higher resolution feature map to locate keypoints' coordinates, we just use the final cell's output in 1/4 scale as the part representation.  Finally, we use $P$ subnetworks (multi-head) sharing backbone to produce $P$ part representations, as shown in Fig.~\ref{framework}. Thus, the local information and pairwise relationships between highly correlated keypoints are combined in each part representation and predicted by each subnetwork. The global information and constraint relationships among all parts are implicitly learned by the shared feature and predicted by the backbone. The long-range and short-range constraint relationships of the human pose are enforced by the whole architecture in an end-to-end learning method.

\subsection{Body Part Representation with Vector in Pixel}
\label{vector in pixel}

Based on the top-down method of pose estimation, we estimate human pose with single person proposal. Given a input image $I\in \mathbb{R}^{H\times W\times 3}$ centered at a person proposal, there will be $P$ part representations $T_1,T_2,...,T_P$ to be predicted. Let $T_p \in \mathbb{R}^{ J \times h^{'}\times w^{'}\times d}$ denote its $p$-th body part representation, where $J$ is the number of keypoints belonging to this part (see the assignment in Tab \ref{group}), i.e. a part what we mean here may associate several keypoints.  $h^{'}$ and $w^{'}$ are the height and width of part representation; in our down-sampling setting, $h^{'}/H=w^{'}/W=1/4$.  $d$ is the dimension of vector in each pixel position. For the $i$-th keypoints of $T_p$, vector in position $\left(x,y\right)$ is denoted as $\vec{\mathbf{v}}_{i,x,y}=T_p^i\left(x,y\right) \in \mathbb{R}^d$, simplified as $\vec{\mathbf{v}}$. Note that the dimension  $d$ of vector is set to 8 by default and choice for dimension is discussed in Section \ref{dimensions} .

We relax scalar value into a latent vector as keypoint entity in each image pixel, expecting it to implicitly capture the redundant and local feature information around the keypoint position, please see more explanation in Appendix~\ref{appendix}. 

Besides inherent to the characteristics of encoding locations, $\vec{\mathbf{v}}$ can represent existing probability of keypoints by using Squashing Function $f_S(\cdot)$ \cite{sabour2017dynamic} to normalize its $\ell_{2}$ norm to $\left[ 0,1\right)$. Formally, for $i$-th keypoint of $p$-th part, the non-linear Squash function will compute the squashed vector $\vec{\mathbf{v}}_s$ in each position $(x,y)$ of $T_p^i$  by:
\begin{equation}
\vec{\mathbf{v}}_s=\frac{\left\|\vec{\mathbf{v}}\right\|^{2}}{1+\left\|\vec{\mathbf{\mathbf{v}}}\right\|^{2}} \frac{\vec{\mathbf{\mathbf{v}}}}{\left\|\vec{\mathbf{\mathbf{v}}}\right\|},
\end{equation}
\begin{equation}
\left\|\vec{\mathbf{v}}_s\right\|=f_{S}(\vec{\mathbf{v}})=\frac{\left\|\vec{\mathbf{v}}\right\|^{2}}{1+\left\|\vec{\mathbf{v}}\right\|^{2}},
\end{equation}
where $\left\|\vec{\mathbf{v}}_s\right\|$ exactly represents $i$-th keypoint's existing score in position $(x,y)$. In inference, the position $(\bar{x},\bar{y})$ of the longest (max-norm) $\vec{\mathbf{v}}$ will be regarded as keypoint location. Predicted score maps are collected from all part presentations. Commonly,  groundtruth score map $H_{k}^{gt}$ is generated from the groundtruth of $k-$th keypoint's position by applying 2D Gaussian with deviation of $\sigma$ where the peak value equals 1 and  $\sigma$ controls the spread of the peak. Train loss $\mathcal{L}_{train}$ is computed by Mean Square Error (MSE) between the predicted score maps and groundtruth score maps for all $P$ part representations. 

Note that the $i$-th keypoint of the part $T_p$ is some type of all keypoints, and several parts may contain the same type of keypoint, e.g. elbow maybe fall into the upper arm part and the lower arm part. Thus we make a indicator function $\mathbbm{1}\left(k,p,i\right)$ equal $1$ if the local index $i$-th keypoint of the part $T_p$ is the global index $k$-th type of all keypoints otherwise $0$. And the final position $(\bar{x},\bar{y})$ will be inferred by the prediction summed from these parts (same effect by averaging). We finally formulate the training loss as:
\begin{equation}
\mathcal{L}_{train} =\frac{1}{K} \sum_{k=1}^K\sum_x \sum_y { \| \hat{H}_k\left( x,y \right) -H_{k}^{gt}\left(x,y \right) \|_2^2 }, 
\end{equation}
\begin{equation}
\hat{H}_k\left( x,y\right) =  \sum_{p,i} f_{S}\left(T_p^i\left( x,y \right) \right)\cdot \mathbbm{1}\left(k,p,i\right),
\end{equation}
where $\hat{H}_k$ is the prediction map for each keypoint, $(x,y)$ is each location of $T_p^i$, $\hat{H}_k$ and $H^{gt}_k$. In practice, we will mask out the keypoints without position annotation in the training process.

\subsection{Prior Knowledge of human body structure}
\label{prior}
The intuition behind this work is that using part-specific neural networks to separately capture each pairwise spatial relationship in local part is possibly better than using a shared neural network to model the global relationship. Therefore, we consider the human body structures and adopt four types of grouping strategy to make comparison. Specially, $P=1$ means that we model long-range dependencies relationships of pose and global relationship of all joints is learned  by a shared neural network. $P=3$ means that body pose is predefined into 3 parts associated with 3 CNFs: head part, upper limb part and lower limb part. $P=8$ means that body pose is predefined into 8 parts associated with 8 CNFs: head-shoulder, left upper arm, left lower arm, right upper arm, right lower arm, thigh, left lower leg and right lower leg. In addition, we also adopt the data-driven grouping strategy of \cite{Tang_2019_CVPR} by setting $P=5$: head-shoulder, left lower arm, right lower arm, thigh, lower limb part. In Tab~\ref{group}, all skeleton keypoints are associated with the corresponding part according to the prior body structure.\footnote{For MPII \cite{andriluka20142d} dataset, we set indices of head top, upper neck, thorax, l-shoulder, r-shoulder, l-elbow, r-elbow, l-wrist, r-wrist, l-hip, r-hip, l-knee, r-knee, l-ankle, r-ankle, pelvis keypoints to 0-15 orderly. For COCO \cite{lin2014microsoft} dataset, we set indices of nose, l-eye, r-eye, l-ear, r-ear, l-shoulder, r-shoulder, l-elbow, r-elbow, l-wrist, r-wrist, l-hip, r-hip, l-knee, r-knee, l-ankle and r-ankle to 0-16 orderly. }

\begin{table}[h]
	\centering
	\caption{According to the prior knowledge of human body structure, there are different grouping types of body part representations.}
	\label{group}
	\renewcommand{\arraystretch}{1.1}
	\begin{tabular}{cccc}
		\toprule
		\multirow{2}{*}{\textbf{Representation Mode}}& 
		\multirow{2}{*}{\textbf{Group Name}}&\multicolumn{2}{c}{\textbf{Index}}\\
	
		&&MPII&COCO\\
		\hline
		$P=1$&all keypoints&0-15&0-16\\
		\hline
		
		\multirow{3}{*}{$P=3$}& head part& 0-2  & 0-4\\
		
		& upper limb part&3-8 &5-10\\
		& lower limb part&9-15 &11-16\\
		\hline
		\multirow{5}{*}{$P=5$}& head-shoulder& 0-4&0-6\\
		&left lower arm&5,7&7,9\\
		&right lower arm &6,8&8,10\\
		&thigh&9,10,15&11,12\\
		& lower limb part&11-14 &13-16\\
		\hline
		
		\multirow{8}{*}{$P=8$}&head-shoulder&0-4&0-6\\
		&left upper arm&3,5&5,7\\
		&left lower arm&5,7&7,9\\
		&right upper arm&4,6&6,8\\
		
		&right lower arm &6,8&8,10\\
		&thigh&9-12,15&11-14\\
		&left lower leg &11,13&13,15\\
		&right lower leg&12,14&14,16\\
		\bottomrule
		
	\end{tabular}

\end{table}
\subsection{Optimization}
\label{optimization}
\textbf{One-shot Search and Part-specific architectures for Different Parts.} Given a hyperparameter $\Theta$ for a CNF, the weights $w_o=\left\{w_{s, l} \right\}$ and architecture $\alpha_o=\left\{\alpha_{s, l} \right\}, \beta_o=\left\{\beta_{s,l} \right\} $ are optimized. Following the principle of one-shot architecture search\cite{elsken2018neural}, we assume that  $\alpha_{s,l}$ share same weights across a single CNF and $\beta_{s,l} $ is cell-wise in each CNF. Supposing that there are $P$ CNFs, their weights of operations are $w=\left\{w_0,...,w_P \right\}$ and the total architecture parameters are $\alpha=\left\{\alpha_1,...,\alpha_P \right\}, \beta=\left\{\beta_1,...,\beta_P \right\}$. We search for part-specific CNF to adapt each prediction sub-task, which means that the architecture parameters $\alpha_1,...,\alpha_P$ are totally distinct and so are $\beta_1,...,\beta_P$. In the Section~\ref{part_modes}, we make contrastive experiments to study the differences on different keypoints groups strategies.

\textbf{Random Sampling.} Random search can be seen as a powerful baseline for neural architecture search or hyperparameter optimization \cite{Xie_2019_ICCV,bergstra2012random,li2019random}. It is conducted in \cite{liu2018darts} as well and has a competitive result compared with the gradient-based method. In this work, we randomly initialize values of $\alpha,\beta$ by standard normal distribution and make them fixed in the whole training process.   From another point of view, this also can be viewed as a \emph{stochastic} network generator like \cite{Xie_2019_ICCV}. The sampled architecture parameters make no assumption about the structures, and only the weights of neural networks are optimized. Therefore, we conduct it to validate the design of CNF, each random experiment also can represent the performance of CNF without any neural architecture search strategies.

\textbf{Synchronous Optimization.} In DARTS \cite{liu2018darts}, the architecture search problem is regarded as a bilevel optimization problem. An extra subset $val$ of original train set is held out serving as performance validation to produce the gradient w.r.t. architecture parameters $\alpha,\beta$ excluding the weights $w$. However, the training of the second-order gradient-based method is still time-consuming and restricted by GPU memory due to the high resolution intermediate representation for pose estimation. Moreover, the training for parent continuous network and the derived net are inconsistent  in DARTS, final pruned network needs to be trained again with all training samples.  There are works attempting to eliminate this problem, such as SNAS \cite{xie2018snas} introducing Gumbel-Softmax trick. Instead, we explore a more simple way as our major optimization strategy, benefit from the parametric form of cell. The $\alpha $ and $\beta$ are registered to model's parameters, synchronously optimized with the weights $w$, i.e. the $w$, $\alpha $ and $\beta$ are updated by $\nabla_{w,\alpha,\beta}\mathcal{L}_{train}$ in a single step of gradient descent. Without extra validation phase, the final continuous $\alpha,\beta$ and the weights $w$ have seen all training samples, thus it do not need to be pruned into discrete architecture by $argmax$ operation and trained again from scratch.

\section{Experiments}

A series of experiments have been conducted on two datasets MPII Human Pose Dataset \cite{andriluka20142d} and COCO Keypoint \cite{lin2014microsoft}. In Section~\ref{implement} and Section \ref{ablation}, we show the implementation details and ablation study for the effectiveness of the purposed optimization strategies, part-based pose representation and the vector in pixel method. Then we compare our model with a part-based baseline model and an efficient light-weight model, in Section \ref{compare}. Finally, in Section \ref{state-of-the-art}, we present the results compared with the state-of-the-art.

\subsection{Implementation Details}
\label{implement}

As for most subnetworks in ablation study, we set $C$=10 and total final layers is 3 (discarding first 3 layers of CNF architecture with $L=6$), and the total number of cells is 6 as a basic configuration. Backbone with fabric can be constructed as described in Section~\ref{subnet}. To make fair comparison with methods using model pretrained on ImageNet \cite{russakovsky2015imagenet}, we take Fabric-1, Fabric-2, Fabric-3, pretrained Mobilenet-V2 \cite{Sandler2018MobileNetV2IR}, ResNet-50 \cite{he2016deep} and HRNet-W32-Stem-Stage3 \cite{Sun_2019_CVPR} feature blocks (5.8M, 6.8M, 10.5M, 1.3M, 23.5M, 8.1M parameters respectively) as choices for backbone to provide feature pyramid to subnetworks, see configuration details in Appendix \ref{cnf_backbone} and \ref{mobilenetv2_backbone}.

We implement our work by PyTorch \cite{paszke2017automatic} and each experiment is conducted on a single NVIDIA Titan Xp GPU. Training epoch is 200 and batchsize is set to 24 (not fixed). We use Adam \cite{kingma2014adam} optimizer to update the weights and architecture parameters with 0.001 initial learning rate, decay at epoch 90, 120, 150 with 0.25 factor by default. Data augmentation strategies are used with random rotation range in $\left[-45^{\circ}, 45^{\circ}\right]$, random scale range in $\left[0.7, 1.3\right]$ and random flipping with 0.5 probability. Flip test is used in inference. In practice, one quarter offset from the peak to the secondary peak is introduced to reduce the quantization error. Strategies mentioned above are adopted in all ablation experiments. 

\subsection{Ablation Study}

\label{ablation}

\textbf{Dataset and Evaluation.} We conduct ablation study on MPII Human Pose Dataset \cite{andriluka20142d} which is a benchmark for evaluation of pose estimation. The dataset consist of around 25K images containing over 40K people with annotated body joints. All models in ablation study experiments are trained on a subset MPII training set and evaluate on a held validation set of 2958 images following \cite{xiao2018simple}. The standard PCKh metric (head-normalized probability of correct keypoint) is used for MPII. PCKh@0.5 means that a predicted joint is correct if its position is within 50$\%$ of the length groudtruth head box from its groundtruth location. Evaluation procedure reports the PCKh@0.5 of head, shoulder, elbow, wrist, hip, knee, ankle, mean PCKh@0.5 and mean PCKh@0.1.

\begin{table}[h]
	\centering
	\caption{Optimization Strategies (search method). We choose MobileNet-V2 as the backbone of model. $P=3,H=1,C=10,d=8$, each subnetwork has six cells and total parameters of model is 3.3M, the Madds of model inference complexity for single input sample is 1.2 GFLOPs.}
	\label{strategy}
	\renewcommand{\arraystretch}{1.2}
	\setlength{\tabcolsep}{0.8mm}{
		\begin{tabular}{cccc}
			\toprule
			\textbf{Search Method}&\textbf{Search Time } &\textbf{Mean}&\textbf{Mean}\\
			
			(search strategy)&(GPU days)&(PCKh@0.5)&(PCKh@0.1)\\
			
			\hline
			Random &0.8&87.1$\pm$0.2&35.2$\pm$0.4\\
			
			First-order gradient-based&0.9&87.1&34.7\\
			
			Synchronous gradient-based&0.8&87.0& 34.6\\
			\midrule
	\end{tabular}}
\end{table}

\begin{table}[h]
	\centering
	\caption{Body Part Representation Modes. We choose MobileNet-V2 as the backbone of model. $H=1,C=10,d=8$, each subnetwork has six cells. }
	\label{modes}
	\renewcommand{\arraystretch}{1.2}
	\begin{tabular}{ccc}
		\toprule
		\textbf{Representation Mode}& 
		\textbf{Mean}(PCKh@0.5)&\textbf{Mean}(PCKh@0.1)\\
		\hline
		$P=1$& 86.4&33.4\\ 
		
		$P=3$&87.0&34.6\\ 
		
		$P=5$&87.1&35.1\\
		
		$P=8$&87.3&35.6\\
		\bottomrule
		
	\end{tabular}
	
\end{table}

\begin{table} \scriptsize
	
	\caption{Dimension Choices for Vector in Pixel. We choose MobileNet-V2 as the backbone of model. $P=3,H=1,C=10$, each subnetwork has six cells and total parameters of model is 3.3M, the Madds (FLOPs) of model inference complexity for single input sample is 1.2 GFLOPs. $\dagger$ is the result achieved by running the official code on our machine, the official is $90.3$.}
	\label{dimensions}
	\centering
	\renewcommand{\arraystretch}{1.2}
	\setlength{\tabcolsep}{0.5mm}{
		\begin{tabular}{ccccc}
			\toprule
			\textbf{Dimension}& \textbf{Mean}(PCKh@0.5)&\textbf{Mean}(PCKh@0.1)&\#\textbf{Params}&\#\textbf{Madds}(FLOPs)\\
			\hline
			$d=1(scalar)$ &86.8&33.5&3.3M&1.1G\\
			$d=4$& 86.9&34.9&3.3M&1.1G\\
			
			$d=8$&87.0&34.6&3.3M&1.2G\\
			
			$d=16$&86.8& 34.9&3.3M&1.2G\\
			
			\midrule
			SimpleBaseline \cite{xiao2018simple}&88.5&33.9&34.0M&12.0G\\
			\quad+ vector(8-dim)&88.7&34.2&34.0M&12.1G\\
			\hline
			HRNet \cite{Sun_2019_CVPR}&90.1$\dagger$&37.7&28.5M&9.5G\\
			\quad+ vector(8-dim)&90.2&38.1&28.5M&9.6G\\
			\toprule

	\end{tabular}}

\end{table}

\begin{table*} \normalsize
	\centering
	\caption{Comparisons of performance, model parameters and inference complexity on MPII test set. The backbones of ours-a and ours-b models are MobileNet-V2 (1.3M) and HRNet-W32-Stem$\sim$stage3 (8.1M).}
	\label{mpii-test}
	\renewcommand{\arraystretch}{1}
	\setlength{\tabcolsep}{1.0mm}{
		\begin{tabular}{ccccccccccc}
			\toprule
			\textbf{Method}&Head&Shoulder & Elbow &Wrist&Hip&Knee&Ankle&Total  &\#Params&\#FLOPs\\
			\hline
			Tompson et al. \cite{tompson2014joint}&95.8 &	90.3 &	80.5 &	74.3 &	77.6 &	69.7 &	62.8& 	79.6&-&-\\
			Belagiannis \& Zisserman \cite{belagiannis2017recurrent} &97.7&95.0&88.2&83.0&87.9&82.6&78.4&88.1&-&-\\
			Wei et al. \cite{wei2016convolutional}&97.8&95.0&88.7&84.0&88.4&82.8&79.4&88.5&-&-\\
			Insafutdinov et al. \cite{insafutdinov2016eccv}& 96.8  & 95.2  & 89.3  & 84.4  & 88.4  & 83.4 & 78.0 & 88.5 &42.6M&41.2G \\
			Bulat\&Tzimiropoulos \cite{bulat2016human}&97.9 &	95.1 &	89.9 &	85.3 &	89.4 &	85.7 &	81.7 &	89.7&-&-\\
			Newell et al. \cite{newell2016stacked}& 98.2  & 96.3  & 91.2  & 87.1  & 90.1  & 87.4 & 83.6 & 90.9&25.1M &19.1G \\
			Xiao et al. \cite{xiao2018simple} &98.5&96.6&91.9&87.6&91.1&88.1&84.1&91.5&68.6M &20.9G\\
			Tang et al. \cite{tang2018deeply}& 98.4  &96.9 &92.6 & 88.7 &91.8 & 89.4& 86.2&92.3&15.5M&33.6G \\
			\hline
			FPD (Knowledge Distillation) \cite{Zhang2018FastHP} &98.3 &96.4 &91.5& 87.4 &90.9& 87.1& 83.7 &91.1 &3M&9G \\
			\hline
			Ours-a& 97.9  & 95.6  & 90.7  & 86.5  & 89.8  & 86.0 & 81.5 & 90.2 & 5.2M &4.6G\\
			Ours-b& 98.2 & 95.9& 91.5 & 87.6  & 90.1 & 87.3 & 83.2 & 91.0 & 16.4M &9.4G\\
			\bottomrule
		\end{tabular}
	}

\end{table*}

\begin{table*}\small  
	\renewcommand{\arraystretch}{1.3}
	\caption{Comparison with baseline models on MPII validation set. We choose HRNet-W32-stem$\sim$stage3 as the backbone, which outputs shared feature pyramids with four levels fo feature maps. Input size is $256\times256$. Our-c-CNF: $P=3,H=2,C=12$. Our-b-CNF: $P=5,H=1,C=16$. $\mathcal{O}$ includes zero, skip connection, $3\times3$ separable conv, $3\times3$ dilated conv with 2 rate. Here we use a 1 × 1 convolution to first reduce the feature dimension to $W=400/296$ and then $D=8/8$ subsequent residual blocks for the Branchnets of Baseline-1 and Baseline-2. $\dagger$ is the result achieved by running the official code on our machine, the official result is 90.3 AP. $\ddagger$ represents the metric Mean PCKh@0.5 over ten hard joints, reported in \cite{Tang_2019_CVPR}.}
	\label{HRNet+fabrics}
	\centering
	\renewcommand{\arraystretch}{1.32}
	\setlength{\tabcolsep}{0.7mm}{
		\begin{tabular}{ccccccccc}
			\toprule
			
			\textbf{Method}&\textbf{Backbone}&\textbf{Head}&\textbf{NAS}&\textbf{Part-specific}& \textbf{Mean@0.5}&\textbf{Mean@0.1}&\#\textbf{Params}&\#\textbf{FLOPs}\\
			\midrule
			SimpleBaseline \cite{xiao2018simple} &ResNet-152 (52.2M) & DeConvs (16.4M)&\ding{55}&\ding{55}&89.6& 35.0 &68.6M&20.9G\\
			
			HRNet-W32 \cite{Sun_2019_CVPR}&Stem$\sim$stage3 (8.1M)&Stage4(19.7M)&\ding{55}&\ding{55}&90.1$\dagger$&37.7&27.8M&9.5G\\
			Stacked PBNs \cite{Tang_2019_CVPR}& Stacked Hourglass Network & Stacked Branchnets  &  \ding{55} & \Checkmark & 88.14$\ddagger$&-& 26.7M&-\\
		\hline
	
			Baseline-1&Stem$\sim$stage3 (8.1M)& Branchnet $\times$ 3 (4.7M)& \ding{55}& \Checkmark &89.0&38.2&12.8M&23.7G\\
			Baseline-2&Stem$\sim$stage3 (8.1M)& Branchnet $\times$ 5 (4.5M)& \ding{55}& \Checkmark  &89.3&38.7&12.6M&22.8G\\
		
			\hline
			Random partition&Stem$\sim$stage3 (8.1M)& CNF $\times$ 3 (4.8M)&\Checkmark &\Checkmark &81.0&30.5&12.9M&8.3G\\
			\hline
			Ours-c&Stem$\sim$stage3 (8.1M)& CNF $\times$ 3 (4.8M)&\Checkmark &\Checkmark &89.9&39.5&12.9M&8.3G\\
			
			Ours-b&Stem$\sim$stage3 (8.1M)& CNF $\times$ 5 (8.3M)&\Checkmark &\Checkmark &90.1&39.4&16.4M&9.4G\\
			\bottomrule	
			
	\end{tabular}}
	
\end{table*}

\begin{figure}
	
	\centering
	
	\includegraphics[height=6cm, width=9cm]{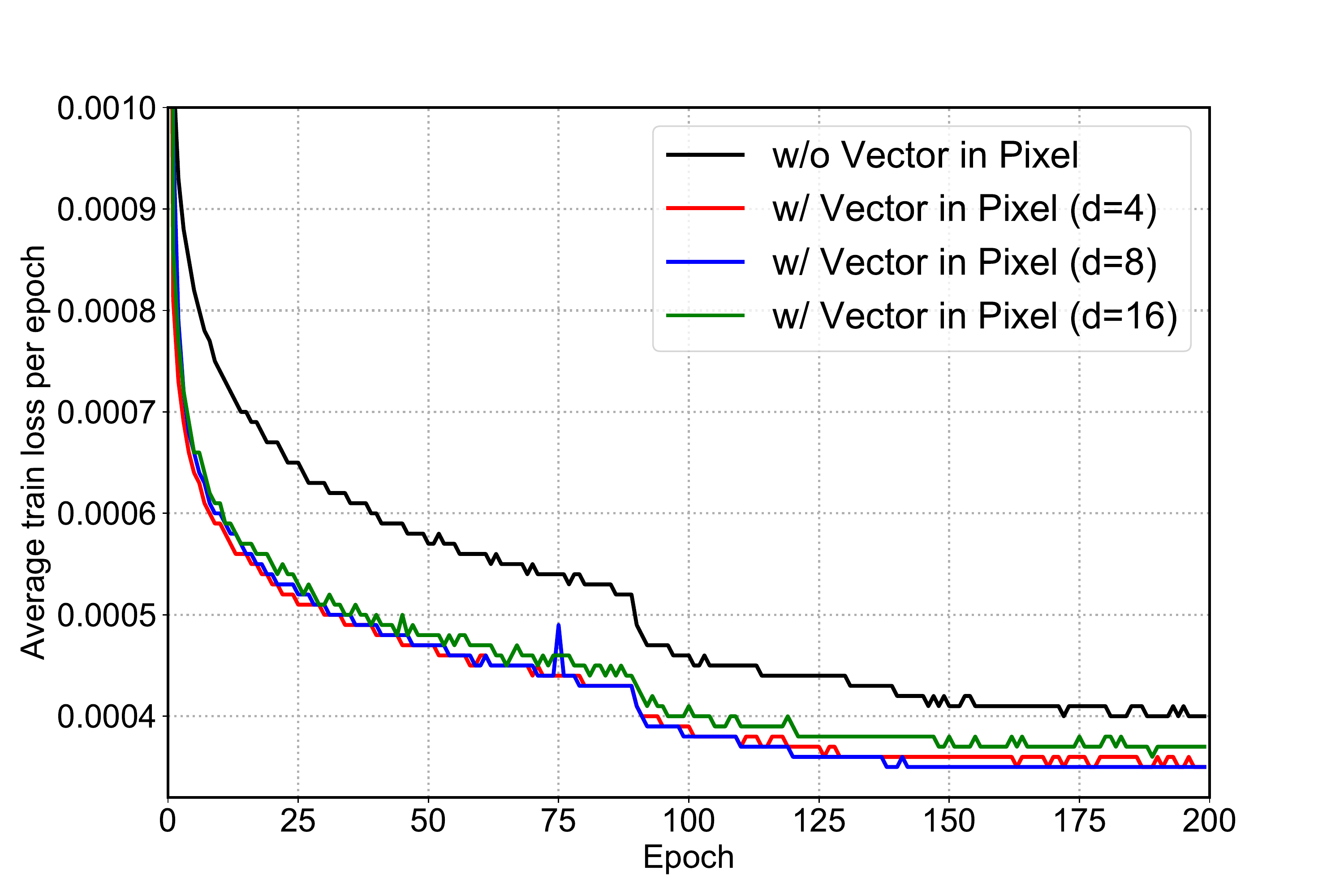}
	\caption{Train losses of experiments w and w/o vector in pixel method. Detailed configurations are described in Tab~\ref{dimensions}. The sudden drop at epoch 90 is caused by learning rate decay.}
	\centering
	\label{w_o_vector_in_pixel}
	
\end{figure}

\begin{figure*}[h]
	
	\centering
	\includegraphics[height=6cm, width=18cm]{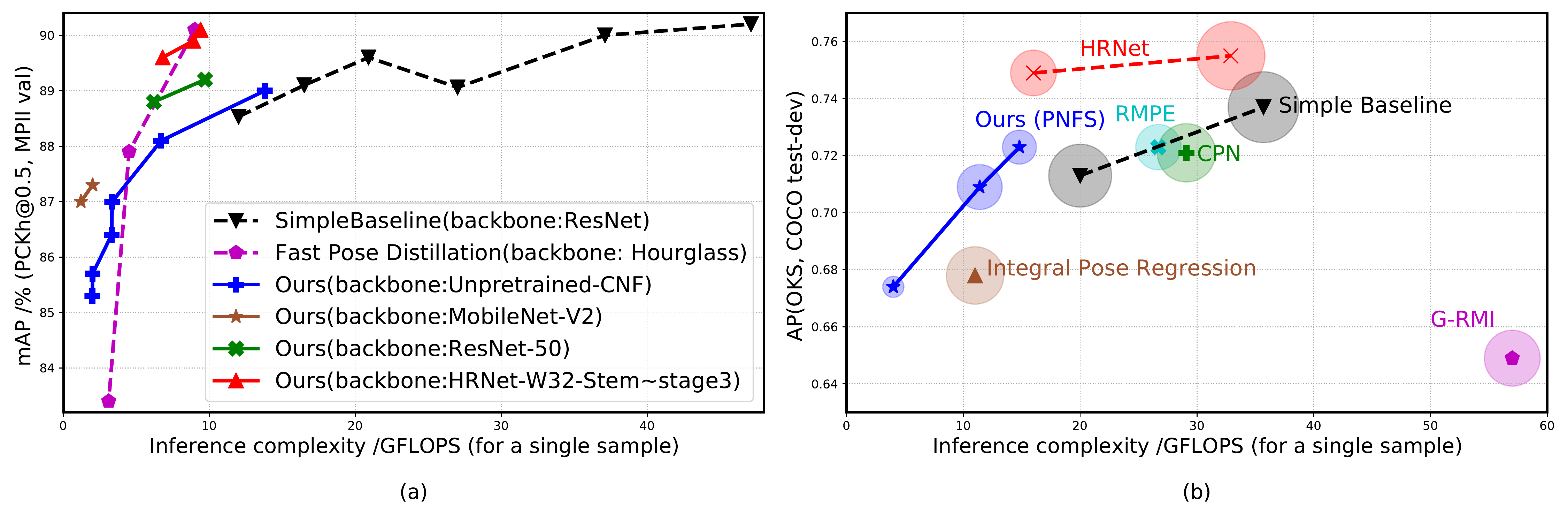}
	\caption{\textbf{(a)}: The mAP of PCKh@0.5 metric vs. model inference complexity (GFLOPs) on MPII val set. \textbf{(b)}: The AP of OKS@0.5:0.95 c vs. model inference complexity (GFLOPs) on COCO test-dev2017 dataset. Model parameters and FLOPs of detecting persons in COCO dataset are not included. The areas of the circles are linearly relative with the amounts of models' parameters. }
	\centering
	\label{MAP-FLOPS}
	
\end{figure*}

\paragraph{Optimization Strategies} For random search strategy, we conduct 5 experiments with different pseudo random seeds, each experiment costs 0.8 days for a single GPU. Result shows that completely random architecture parameters can perform well. As shown in Tab~\ref{strategy}, synchronous optimization is effective as well as random search under the same configuration and search time. We observe that the best result of random initialization for architecture surpasses the synchronous optimization, this actually reveals two points: 1) the search space design has more significant impact on the performance of neural network; 2) the parameter initialization becomes important when not leveraging NAS to the proposed CNF structure. In addition, we implement the first-order gradient-based optimization method according to the official code\footnote{https://github.com/quark0/darts/blob/master/cnn/architect.py} of DARTS \cite{liu2018darts} for comparison. We hold out half of MPII training data as validation for performance estimation of architecture. Another Adam optimizer is used to update $\alpha, \beta$ with 0.003 learning rate and 0.001 weight decay, discrete architecture is not derived from continuous architecture for full training. Note that we can not theoretically prove the found architecture is the optimal one, but we find that there is almost no performance difference between the synchronous optimization and the first-order gradient-based optimization proposed by DARTS, which demonstrates its effectiveness compared with the first-order optimization.

\begin{table*}[h]  \small  
	\renewcommand{\arraystretch}{1.1}
	\caption{Comparisons of performance, model parameters and inference complexity on COCO test-dev set. Model parameters and FLOPs of detecting persons are not included. The backbones of ours-1, ours-2 and ours-3 models are MobileNet-V2 (1.3M), ResNet-50 (23.5M) and HRNet-W32-Stem$\sim$stage3 (8.1M).} 
	\label{coco}
	\centering
	
	\renewcommand{\arraystretch}{1}
	\setlength{\tabcolsep}{0.9mm}{
		\begin{tabular}{ccccccccccccc}
			\toprule
			&$AP$ & $AP^{50}$& $AP^{75}$ &$AP^{M}$  &A$P^{L}$&$ AR$ &$ AR^{50}$ & $AR^{75} $&$AR^{M}$ & $AR^{L}$ & \#Params&\#FLOPs\\
			\hline
			
			CMU-Pose \cite{cao2017realtime} & 0.618&0.849 &0.675&0.571&0.682&-&-&-&-&-&-&-\\
			
			Mask-RCNN \cite{he2017mask}&0.631&0.873&0.687&0.578 &0.714&-&-&-&-&-&-&-\\
			Associative Embedding \cite{newell2017associative}&0.655&0.868&0.723&0.606&0.726 &  0.702&   0.895  & 0.760 &  0.646  & 0.78&-&-\\
			Integral Pose Regression \cite{sun2018integral}&0.678&0.882&0.748&0.639&0.74&-&-&-&-&-&45.0M&11.0G\\
			
			SJTU\cite{fang2017rmpe} &0.680&0.867&0.747&0.633&0.750&0.735&0.908&0.795&0.686&0.804&-&-\\
			G-RMI \cite{papandreou2017towards} &0.685&0.871&0.755&0.658&0.733&0.733&0.901&0.795&0.681&0.804&42.6M&57.0G\\
			PersonLab \cite{Papandreou2018PersonLabPP} &0.687 &0.890&0.754&0.641&0.755&0.754&0.927&0.812&0.697&0.830&-&-\\

			MultiPoseNet \cite{kocabas2018multiposenet} & 0.696&0.863&0.766&0.650&0.763&0.735&0.881&0.795&0.686&0.803&-&-\\
			CPN \cite{chen2018cascaded} & 0.721 &0.914& 0.800& 0.687& 0.772& 0.785&0.951 &0.853 &0.742 &0.843 &-&-\\
			SimpleBaseline(ResNet-50) \cite{xiao2018simple}&0.702&0.909&0.783&0.671&0.759&0.758&-&-&-&-&34M &8.9G\\
			SimpleBaseline(ResNet-152) \cite{xiao2018simple}&	0.737&0.919&0.811&0.703&0.800&0.790&-&-&-&-&68.6M &35.6G\\
			HRNet-W32 \cite{Sun_2019_CVPR}& 0.749& 0.925& 0.828 &0.713& 0.809& 0.801&-&-&-&-& 28.5M &16.0G\\
			\hline
			Ours-1& 0.674& 	0.890 &	0.737 &	0.633 & 	0.743 & 	0.731 & 	0.928& 	0.791 & 	0.681 & 	0.800&6.1M& 4.0G\\
			Ours-2& 0.709& 	0.904 &	0.777 &	0.667 & 0.782 & 	0.766 & 	0.941& 	0.829 & 0.715& 	0.836&27.5M& 11.4G\\
			  Ours-3&  0.723 &  0.909 &  0.795 &	  0.684&  	0.792&	  0.779 &	  0.945&	  0.844&  	0.731 &  	0.845 &   15.8M&   14.8G\\
			\bottomrule
	\end{tabular}}
	
\end{table*}

\paragraph{Body Part Representation Modes}
\label{part_modes}

  We study these four modes predefined by the prior knowledge and results are shown in Tab~\ref{modes}. We choose MobileNet-v2 \cite{Sandler2018MobileNetV2IR} feature blocks as backbone. We find that multiple part presentations predicted by part-specific CNFs surpasses global whole-body representation predicted by a shared CNF. 8 part representations mode achieve 1\% accuracy increase than whole-body representation and mode with $P=3$ is a trade-off between performance and model capacity.

  \paragraph{Dimension Choices for Vector in Pixel}
  
  We study the effect of choice for dimension $d$ of the vector on performance by setting $d$ with $1,4,8,16$. $d=1$ represents the common heatmap regression approach without vector in pixel. We find that 8-dim vector has a better performance shown in Tab~\ref{dimensions}. To validate the generalization of 8-dim vector representation method, we apply it to SimpleBaseline \footnote{https://github.com/microsoft/human-pose-estimation.pytorch} \cite{xiao2018simple} and HRNet \footnote{https://github.com/leoxiaobin/deep-high-resolution-net.pytorch} \cite{Sun_2019_CVPR}. We find that this 8-dim vector representation is effective in these two frameworks. It gains 0.23\% increase in PCKh@0.5 and 0.88\% increase in PCKh@0.1 than SimpleBaseline official results with little increase of complexity and 1.06\% increase in PCKh@0.1 than HRNet official results. Although we find that there is no obvious boost on PCKh@0.5 metric and a little in PCKh@0.1 metric, from Fig.~\ref{w_o_vector_in_pixel} we observe that the losses of training with vector in pixel method converge faster, which implies the fitting between training data and label is more robust. 
  
  \subsection{Comparison with Baseline Model and Efficient Model}
  \label{compare}
  
  \paragraph{Combination and Comparison with Baseline Model}
  \cite{Tang_2019_CVPR} conducts part-specific feature learning for pose estimation as well. It uses a part-based branch network (PBN) that has a shared representation extracted by a Hourglass Network and is stacked by 8 times. It represents our main competitor. For the sake of fairness, we follow the \cite{Tang_2019_CVPR} to construct the Branchnet serving as the subnetworks  but not stack the whole module repeatedly. We choose the ImageNet pretrained HRNet-W32 \cite{Sun_2019_CVPR} as backbone by replacing blocks of the last stage, which only retains 8.1M parameters. And then we make a comparison between NAS-based CNFs and hand-designed Branchnets. By controlling their parameters amount to be similar, we find that using NAS to learn part-specific CNFs has achieved significantly higher AP (89.9 vs. 89.0) and less FLOPs (8.3G vs. 23.7G) than using a fixed structure for each part, as shown in Tab~\ref{HRNet+fabrics}. 
  
  With 16.4M model size and 9.4 GFLOPs, our model achieves 90.1/39.4AP accuracy in PCKh@0.5/PCKh@0.1 metrics, in contrast to 90.3/37.7AP reported from \cite{Sun_2019_CVPR} by HRNet-W32 (28.5M, 9.5G FLOPs) in single-scale testing. We can see that with obviously fewer parameters ($\downarrow$58\%) compared with the stage4 block of HRNet, the searched CNFs can maintain the high performance at PCKh@0.5 metric and improve PCKh@0.1 metric performance significantly.
  
  To validate whether the prior knowledge is significant to help NAS search part-specific neural architectures, we make a contrastive experiment by randomly partitioning keypoints into three groups: 1) l-shouder, l-ankle, l-elbow, pelvis, r-wrist, head-top; 2) upper-neck, r-knee, r-elbow, r-hip, l-wrist, throax; 3) l-hip, r-shoulder, l-knee, r-ankle. Such partitioned groups can be seen as wrong knowledge of human body structure or chaotic training signal. We use the same architecture configurations as Ours-c model, with 12.9M parameters and 9.3 GFLOPs. The best accuracy in the training process is only 81.0 AP and the model collapses at 120 epoch with 41.7 AP, as shown in Tab.~\ref{HRNet+fabrics}. This result indicate that the guidance of correct prior knowledge can help search higher-performances neural architectures, but incorrect targets also make NAS degenerate. We believe that the intervention of appropriate domain knowledge would be significant for NAS to search task-specific architectures.
  
  \paragraph{Comparison with the Efficient Model}
  
  Fast human pose distillation (FPD) \cite{Zhang2018FastHP} is an ideal efficient model to compare. Though it uses knowledge distillation technique rather than NAS, but to obtain the light-weight models is the same goal for both methods. Specifically, it first trains a large Teacher network, Houglass Network, to achieve a high accuracy performance on the task-dataset. Then a small Student model also taking Hourglass network as backbone is trained by knowledge distillation. We show the trade-off curves between AP and FLOPs in the (a) of Fig.~\ref{MAP-FLOPS}. The results show that both methods can achieve competitive performance with less computing complexity for MPII dataset. Our smaller models have a slight advantage over the smaller models of FPD.

\label{prune}
\textbf{Pruning Useless Structures.} In practical, we find that some $\alpha, \beta$ architecture parameters of the final searched architectures might be zero values. That is, cells whose outputs are multiplied by the zero $\beta$ parameter have no computing contributions for the cells in the next layers; operations whose outputs are multiplied by the zero $\alpha$ parameter have no computing contributions for the next hidden node. To further reduce the parameters and computing complexity, we use empty cells and zero operations without parameters to replace those useless cells and operations. By this way, the architectures only retain the structures associated with non-zero architecture parameters. This method does not affect the precision and also does not need to retrain the architecture. Note that our NAS method cannot ensure there are always existing zero values in the micro or macro architecture parameters. When we cancel the learning rate decay of optimizing the architecture parameters so that they remain sensitive to be optimized in the late stage of the search process, we find that some architecture parameters converge to zero values, i.e. some types of operation or cells have no contributions to the forward computing in some searched architectures. We expect that future works can introduce sparse constraint to optimize the architecture parameters, to guarantee the architecture could be pruned by this way.

\begin{figure*}
	
	\centering
	\includegraphics[height=9cm, width=18cm]{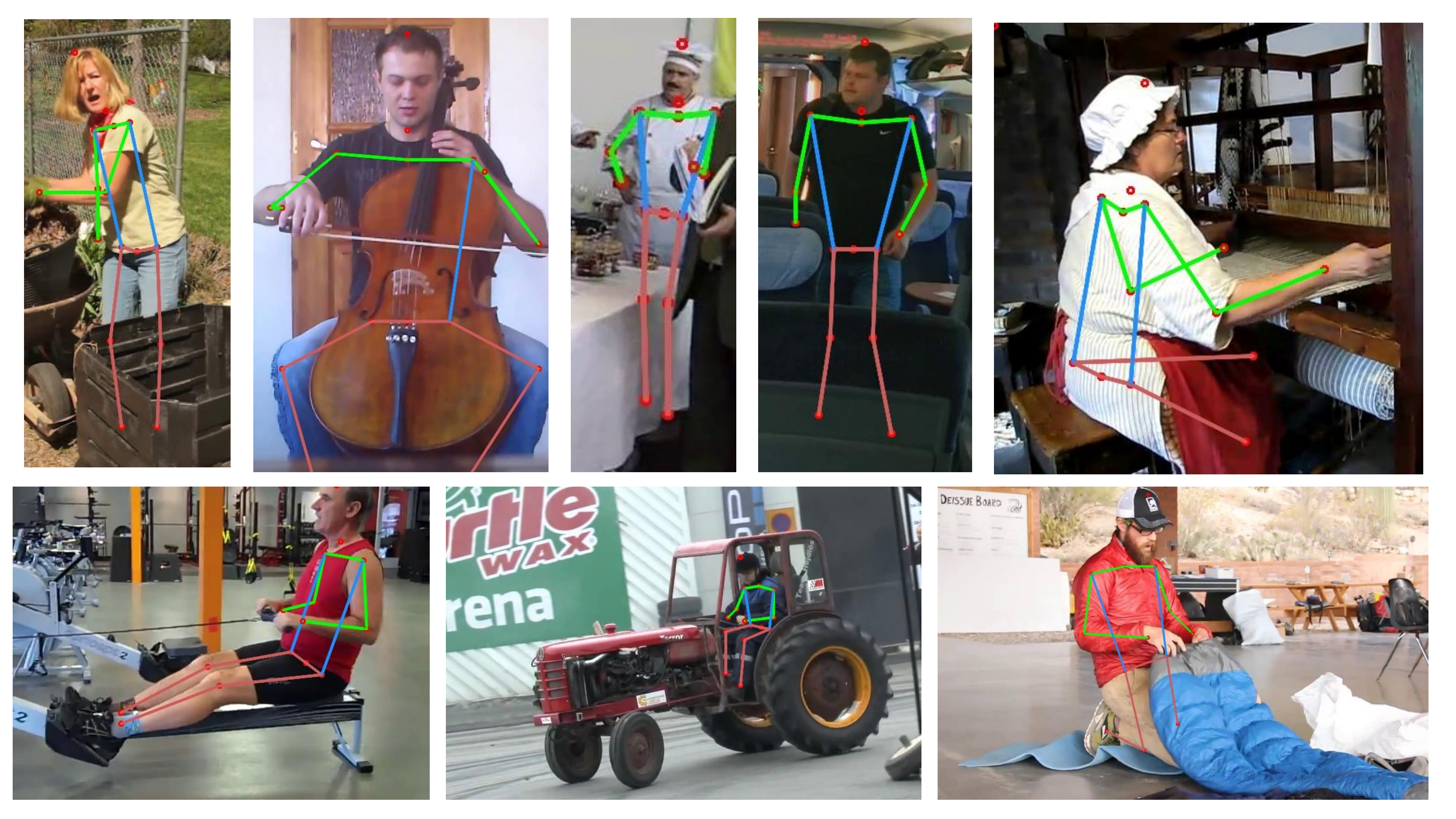}
	\caption{Qualitative pose estimation results on MPII val set for single person pose estimation. We show the cropped image regions containing human body.
	}
	
	\label{mpii_results}
\end{figure*}

\begin{figure*}
	
	\centering
	\includegraphics[height=6.6cm, width=18cm]{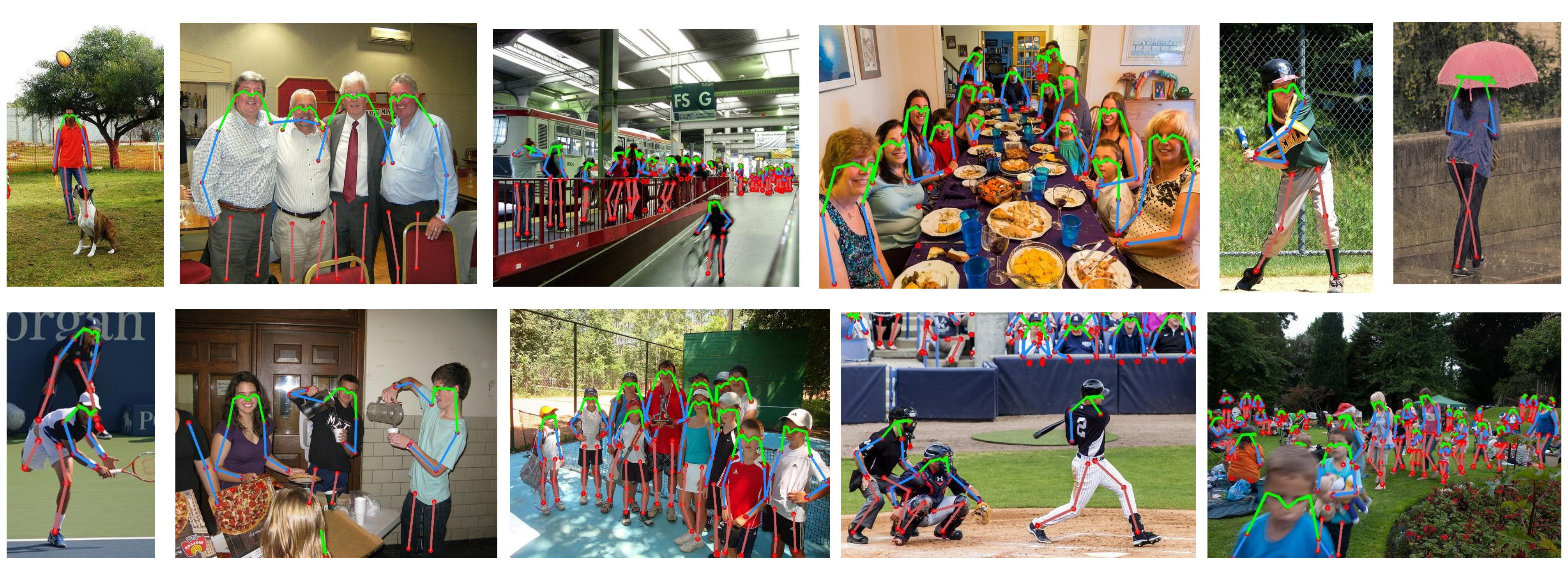}
	\caption{Qualitative pose estimation results on COCO val2017 set. Estimation is conducted on bounding boxes detected by Faster-RCNN \cite{ren2015faster}. It is worth noting that our method works well in some heavily partial occluded hard samples (such as left two images in first row and the fourth in second row).
	}
	
	\label{coco_results}
\end{figure*}

\subsection{Comparison with the state-of-the-art}
\label{state-of-the-art}
\paragraph{Testing on MPII Single Person Pose Estimation} To further evaluate our pose estimation method on test set of MPII \cite{andriluka20142d}, we train the model on all samples of MPII train set with early-stopping strategy. All input images are resized to $384\times384$ pixels, data augmentation is the same as mentioned above. We use pretrained MobileNet-v2 and HRNet-W32-Stem$\sim$stage3 feature blocks as backbone. The hyperparameters are: $P=3/5,H=2/1,d=8/8,C=10/16$ and $\mathcal{O}$ includes zero, skip connection, $3\times3$ separable conv, $3\times3$ dilated conv with 2 rate. The optimization method is synchronous optimization. As shown in Tab~\ref{mpii-test}, we achieve a comparable result (91.0 vs. 91.1 AP and 9.4G vs. 9.0G FLOPs) compared with FPD \cite{Zhang2018FastHP} that exploits knowledge distillation to achieve a lightweight and efficient model.

\paragraph{COCO Keypoint Detection Task} MS-COCO \cite{lin2014microsoft} dataset contains more than 200k images and 250k person instances with keypoints label. We use COCO train2017 as our training set, it consists of 57k images and 150k person instances. Val2017 set contains 5k images and test-dev2017 consists of 20k images. It is worth mentioning that some invisible keypoints are labeled on train set and statistics show that around 11.3 \% of annotated keypoints are invisible according to train2017 annotations. Object keypoint similarity (OKS) is the standard evaluation metric for keypoints locating accuracy. More detailed information is available in COCO official website \footnote{http://cocodataset.org/}.

\label{cocotest}
COCO keypoint detection task involves detecting bodies and localizing their keypoints. Based on top-down method, we focus on single pose estimation, therefore we use the detected bounding boxes detected by Faster-RCNN \cite{ren2015faster} with 60.9 AP persons detection results on COCO test-dev2017 dataset. We respectively use pretrained MobileNet-v2 \cite{Sandler2018MobileNetV2IR}, ResNet-50 \cite{he2016deep} and HRNet-W32-Stem$\sim$stage3 \cite{Sun_2019_CVPR} feature blocks as backbone and train two models only on train2017 set. The hyperparameters are: $P=3/3/5,H=1/1/1,d=8/8/8$ and $\mathcal{O}$ includes zero, skip connection, $3\times3$ separable conv, $3\times3$ dilated conv with 2 rate. $C=16/10/16$ for ours-1/ours-2/ours-3 models. The optimization method is synchronous optimization, we cancel the learning rate decay for searching the architecture parameters of ours-3 model, and prune it as described in Section\ref{prune}. The input size is $384\times288$ pixels and the OKS-NMS algorithm \cite{papandreou2017towards} is utilized to suppress redundant detected bounding boxes . We report average precision (AP) and average recall (AR) on COCO test-dev2017 set.  With fewer parameters and low computational complexity, we can achieve comparable results with some state-of-the-art methods without using any extra data, ensemble models or other training tricks, 76.2 AP (gt bbox)  and 73.0 AP (detected bbox) on COCO validation set and 72.3 AP on COCO test-dev2017 set, as shown in Tab~\ref{coco}. This model runs at $\sim$20 FPS on a single GPU. The tendency of the trade-off curves between the average precision (AP) and computational cost (FLOPs) shown in the (b) of Fig.~\ref{MAP-FLOPS}. can illustrate our model can achieve comparable performances with many other state-of-the-art methods \cite{fang2017rmpe, xiao2018simple, chen2018cascaded}. Prediction results on some partial occluded hard samples can be seen in Fig.~\ref{coco_results}.

In fact, our searched models are still over-parameterized, the sizes of them also have a potential space to be further reduced by pruning and other model compressing techniques. Large-capacity models (e.g. $>$30M, $>$20GFLOPs) for COCO dataset are not searched, limited by its huge resource consumption caused by the gradient-based NAS search strategy. More advanced neural search strategies can be explored to overcome this problem in the future.

\section{Conclusion}

In this work, we presented a new paradigm - part-specific neural architecture search for human pose estimation, in which we made the first attempt to exploit prior knowledge of human body structure to search part-specific neural architectures automatically. Such paradigm develops the part-based method and gives a new example as Multi-task Neural Architecture Search. Experiment results showed that our light-weight models achieved comparable results on MPII dataset with fewer parameters and lower computational complexity than some state-of-the-art methods. For more challenging COCO keypoint detection task, our light-weight model attained comparable results to some state-of-the-art methods with fewer parameters. In addition, we empirically demonstrate the effectiveness of representing the human body keypoints as vector entities at image locations.  We hope that these ideas may be helpful to adequately leverage NAS to practical application for human pose estimation task or other domains.

\ifCLASSOPTIONcaptionsoff
  \newpage
\fi

\bibliographystyle{unsrt} 
\bibliography{ref}

\appendices

\section{Explanation for the relationship between expected $\left\|\vec{\mathbf{v}}\right\|$ and supervision level $p$}
\label{appendix}
In the section~\ref{vector in pixel}, we use Squash Function to normalize $\vec{\mathbf{v}}$ to $\vec{\mathbf{v}}_s$ whose length ranges in $[0, 1)$,
\begin{equation}
\vec{\mathbf{v}}_s=\frac{\left\|\vec{\mathbf{v}}\right\|^{2}}{1+\left\|\vec{\mathbf{v}}\right\|^{2}} \frac{\vec{\mathbf{v}}}{\left\|\vec{\mathbf{v}}\right\|},
\end{equation}
\begin{equation}
\left\|\vec{\mathbf{v}}_s\right\|=\frac{\left\|\vec{\mathbf{v}}\right\|^{2}}{1+\left\|\vec{\mathbf{v}}\right\|^{2}},
\end{equation}
\begin{equation}
\left\|\vec{\mathbf{v}}\right\|=\sqrt{\frac{1}{1-\left\|\vec{\mathbf{v}}_s\right\|}-1},
\end{equation}
where $\left\|\vec{\mathbf{v}}_s\right\|$ is supervised by numerical value p in each pixel position from groundtruth score maps. Ideal value of $\left\|\vec{\mathbf{v}}_s\right\|$  , denoted as $\left\| \vec{\mathbf{v}}_*\right\|$, equals to $c\in [0, 1)$. Therefore, $\left\|\vec{\mathbf{v}}\right\|$ is supervised by $\sqrt{\frac{1}{1-c}-1}$.

\textbf{Intuition and Explanation.} The extra advantage of vector in pixel is that ambiguity between image feature and groundtruth position can be reduced in some cases of occlusion. In supervised learning, the difficulty of fitting label is usually not under consideration, hard or easy samples of the same category receive the same level of supervision. This issue occurs in keypoints localization because image appearance varies dramatically in some partially occluded and non-occluded areas. Once the area within the keypoint groundtruth position is occluded, the image feature around the keypoint will be disturbed, (e.g. the first image in the~Fig. \ref{coco_results}, the man's ankle is occluded by a dog, but his ankle's position is labeled), as a result it becomes hard to force the network to predict high confidence to match the strong supervision. In such case, our method can handle it as $\left\|\vec{\mathbf{v}}_s\right\|$ replaces $\left\|\vec{\mathbf{v}}\right\|$ under supervision (element value of each dimension of vector has no explicit property and is unsupervised) and the expected length for $\vec{\mathbf{v}}$ in groundtruth keypoint pixel is not directly supervised by numerical value $c$ from groundtruth score but supervised by $\sqrt{\frac{1}{1-c}-1} \in\left[ 0,+\infty\right)$ where $ c\in\left[ 0,1\right) $. In a slight abuse of notation, we write $\left\| \vec{\mathbf{v}}_*\right\| $ as the expected length of $\vec{\mathbf{v}}$, which provides a relatively loose range space for $\vec{\mathbf{v}}$, even if under strong supervision.  

\section{MoblieNet-V2 Backbone Architecture Details}

\subsection{CNF Backbone Architecture Details}
\begin{table}[h]
	\renewcommand{\arraystretch}{1.3}
	\label{cnf_backbone}
	\caption{The detailed configurations for Fabric-1,2,3. The Layers reserved means the number of layers reserved by discarding the in the latter layers of CNF}
\centering

\begin{tabular}{c|c|c|c}
	\toprule[0.2em]
	
- &Fabric-1&Fabric-2&Fabric-3\\
\toprule[0.2em]
L&	7&8& 8\\
C& 10& 10&12\\
H& 2 & 1 & 1\\
Layers reserved& 3&5&5\\
Number of Cells  & 9 &17&17\\

	\toprule[0.2em]
\end{tabular}
\end{table}

\subsection{MoblieNet-V2 Backbone Architecture Details}
\begin{table}[h]
	\label{mobilenetv2_backbone}
	\caption{Given a $H\times W\times 3$ RGB image, The layer of MobileNet-V2 backbone. Each line describes a sequence
		of 1 or more identical (modulo stride) layers, repeated
		n times. All layers in the same sequence have the same
		number c of output channels. The first layer of each
		sequence has a stride s and all others use stride 1. The expansion
		factor t is always applied to the input size. P1$\sim$P4 are taken as the feature pyramids that are send to each CNF. See more configurations in the paper \cite{Sandler2018MobileNetV2IR}}
	\centering
	\begin{tabular}{c|c|c|c|c|c}
		\toprule[0.2em]
		Input Size & Operator                           & $t$& $c$ & $n$ & $s$\\
		\toprule[0.2em]
		$H\times W \times 3$ &    conv2d                  					&  - &  								32 & 1 & 2\\
		$\frac{H}{2} \times \frac{W}{2} \times 32  $ 						&    bottleneck    		&  1 & 16   &1 & 1\\
		$\frac{H}{2} \times \frac{W}{2}\times16$ 							&   bottleneck    		 		&  6 & 24   &2 &2\\
		$\frac{H}{4} \times \frac{W}{4}\times24$ $\rightarrow$ P1&   bottleneck     		   &  6 & 32   & 3    & 2\\
		$\frac{H}{8} \times \frac{W}{8}\times32$ 							& bottleneck       			 &  6 & 64   &  4 &2 \\
		$\frac{H}{8} \times \frac{W}{8}\times64$ $\rightarrow$ P2&    bottleneck 				&  6 & 96   &3 & 1 \\
		$\frac{H}{16} \times \frac{W}{16}\times96$ $\rightarrow$ P3&    bottleneck    		  &  6 & 160  &3  & 2 \\
		$\frac{H}{32} \times \frac{W}{32}\times160$ $\rightarrow$ P4&    -        &  - & -  & - & - \\

		\toprule[0.2em]
	\end{tabular}
\end{table}
%

%
%
%




\end{document}